\documentclass[conference]{IEEEtran}
\IEEEoverridecommandlockouts

\usepackage{cite}
\usepackage{float}
\usepackage{url}
\usepackage[dvipsnames,svgnames]{xcolor}
\usepackage{adjustbox}

\usepackage[framemethod=TikZ]{mdframed}
\usepackage{amsmath,amssymb,amsfonts}
\usepackage{algorithmic}
\usepackage{graphicx}
\usepackage{pifont}
\usepackage{makecell}
\usepackage[colorlinks=true,urlcolor=Blue!90,linkcolor=Blue, citecolor=Blue]{hyperref}
\usepackage{balance}
\usepackage{tabularx} 
\usepackage{array}
\usepackage{multirow}

\def\BibTeX{{\rm B\kern-.05em{\sc i\kern-.025em b}\kern-.08em
    T\kern-.1667em\lower.7ex\hbox{E}\kern-.125emX}}
\begin{document}

\title{\texttt{SentimentGPT}: Exploiting GPT for Advanced Sentiment Analysis and its Departure from Current Machine Learning}

\author{\IEEEauthorblockN{Kiana Kheiri}
\IEEEauthorblockA{\href{https://dsa.cs.usu.edu}{Data Science and Applications lab}  \\
Utah State University\\
Logan, UT, USA \\
\href{mailto:kiana.kheiri@usu.edu}{kiana.kheiri@usu.edu} }
\and
\IEEEauthorblockN{Hamid Karimi}
\IEEEauthorblockA{\href{https://dsa.cs.usu.edu}{Data Science and Applications lab} \\
Utah State University\\
Logan, UT, USA   \\
\href{mailto:hamid.karimi@usu.edu}{hamid.karimi@usu.edu}
}}

\maketitle

\begin{abstract}
This study presents a thorough examination of various Generative Pretrained Transformer (GPT) methodologies in sentiment analysis, specifically in the context of Task 4 on the SemEval 2017 dataset. Three primary strategies are employed: 1) prompt engineering using the advanced GPT-3.5 Turbo, 2) fine-tuning GPT models, and 3) an inventive approach to embedding classification. The research yields detailed comparative insights among these strategies and individual GPT models, revealing their unique strengths and potential limitations. Additionally, the study compares these GPT-based methodologies with other current, high-performing models previously used with the same dataset. The results illustrate the significant superiority of the GPT approaches in terms of predictive performance, more than 22\% in F1-score compared to the state-of-the-art. Further, the paper sheds light on common challenges in sentiment analysis tasks, such as understanding context and detecting sarcasm. It underscores the enhanced capabilities of the GPT models to effectively handle these complexities. Taken together, these findings highlight the promising potential of GPT models in sentiment analysis, setting the stage for future research in this field.\footnote{The code can be found at \url{https://github.com/DSAatUSU/SentimentGPT}}
\end{abstract}


\section{Introduction}

The advent of the digital era has marked an exponential surge in user-generated content due to the rapid advancement and widespread adoption of web technologies ~\cite{karimi2018end,vandam2018cadet}. This burgeoning digital landscape has given the public a diverse range of platforms to freely express their thoughts and opinions and discuss any topic of interest, thereby transforming traditional communication channels. Such a content-rich environment, dominated by social media platforms, has proven to be a substantial source of behavioral insights in different fields such as politics~\cite{karimi2018multi,brookhouse2021road,karimi2019learning,karimi2019multi}, market research~\cite{patino2012social}, education~\cite{karimi2019roadmap,karimi2020towards,solanki2023leveraging,karimi2020understanding,karimi2023analysis,knake2021educational,karimi2021automatic,karimi2022teachers}, humanitarian responses~\cite{reuter2020social}, cybersecuity~\cite{karimi2018end,vandam2018cadet}, healthcare~\cite{braghieri2022social}, urban planning~\cite{lin2019can}, and so on. Leveraging these insights has become increasingly vital in shaping strategies and policies in these diverse domains.  This vast digital space, fueled by user-generated content, offers significant insights, which, when harnessed correctly, can improve decision-making processes and drive quality outcomes.

However, the sheer volume of such data presents an arduous task for manual processing due to human limitations. Sentiment analysis, a vibrant research area within Natural Language Processing (NLP), emerges as a solution to this issue, offering an automatic mechanism to analyze people’s opinions, emotions, and attitudes~\cite{xing2019cognitive}. Sentiment analysis has grown significantly over the past decade, primarily within machine learning, facilitating the extraction, identification, and characterization of sentiments embedded within textual data~\cite{Tsapatsoulis2019Opinion,Balahur2015Sentiment}. Its power to comprehend the attitudes, opinions, and emotions expressed within text data extends beyond theoretical realms~\cite{Wang2017Generative,Biondi2016Web}. Real-life applications of sentiment analysis have surged, with organizations employing it for a variety of purposes, such as monitoring brand perception~\cite{Axhiu2014The}, analyzing political campaigns~\cite{Xia2021Tweet}, enhancing customer service~\cite{Younis2015Sentiment}, and gaining consumer insights~\cite{Bose2019Sentiment}, sentiment analysis is revolutionizing decision-making and strategic planning across industries.
Further expanding the applications, sentiment analysis is also revolutionizing sectors like healthcare and education. Within healthcare, sentiment analysis can be instrumental in understanding patient sentiments towards their treatments and experiences in hospitals~\cite{Aattouchi2021Sentiment,Zhang2021Home}. By interpreting patient feedback on different health policies, caregivers can gain invaluable insights, thereby enabling improved patient care and policy formulation~\cite{Liu2021Public}.
The education sector, too, is harnessing the power of sentiment analysis. Institutions can analyze students' feedback on courses and teachers to gauge the overall sentiment, which can significantly influence their curriculum development and teaching strategies. This approach offers a data-driven way to enhance the quality of education and foster institutional growth~\cite{Santhi2021Sentiment,Sharma2021Stress}.

Nevertheless, despite the substantial strides made in the field, sentiment analysis grapples with challenges primarily stemming from the complex nature of human language and its interpretation~\cite{Sudhir2021Comparative}. This is especially the case when dealing with sentiment analysis of social media content such as Twitter, where the content is usually short~\cite{wankhade2022survey} and noisy~\cite{Omuya2021Sentiment}. More specifically, some notable hurdles are understanding the context of language, detecting sarcasm and irony, deciphering sentiment-bearing linguistic nuances, and grappling with linguistic constructs such as emojis or abbreviations.

In this context, to address these challenges and achieve an accurate sentiment prediction model, we argue that Large Language Models (LLMs), especially GPT, can come to the rescue. 
Recently, LLMs have emerged as a promising solution to many NLP tasks, e.g., paraphrasing~\cite{witteveen2019paraphrasing}, machine translation~\cite{zhang2023prompting}, adversarial training~\cite{liu2020adversarial}, and watermarking task~\cite{kirchenbauer2023watermark}. With their unparalleled size and scale, LLMs can process and comprehend various linguistic constructs and contexts. Their extensive training in diverse corpora of text enables them to capture the intricate nuances of language, including the complex aspects of sentiment, context, and idiomatic expressions~\cite{chen2021evaluating}. Furthermore, their generative capabilities allow them to produce text that maintains coherence and relevance over extended passages~\cite{dathathri2019plug}, a critical factor in tasks like sentiment analysis. With these advantages, LLMs stand as a beacon of hope in surmounting the prevailing challenges in sentiment analysis. More specifically, modern language models like GPT3, boasting an impressive 175 billion parameters trained on 45TB of text, have displayed extraordinary aptitude in numerous language tasks with minimal or no data~\cite{floridi2020gpt}. GPT3 has evolved into GPT3.5, further refining its abilities through code pretraining, as seen in Codex~\cite{siddiq2023exploring}, and reinforcement learning-based fine-tuning instructions. The resultant AI model, ChatGPT, launched in late 2022, quickly gained attention with its strikingly human-like language comprehension and generation skills. In response to these advancements, this paper proposes to explore the potential of GPT for sentiment analysis. Unlike current sentiment analysis modeling, the recent trend in NLP research, known as prompting, employs a novel tactic where a preset textual task description or prompt guides the language model to execute the task accurately~\cite{wei2022chain,fu2022complexity}. This tactic is supplemented by an in-context learning approach.

This study centers around three critical research questions: 
\begin{enumerate}

    \item [\ding{113}]\textbf{RQ1:} How do GPT-related models perform on sentiment analysis of social media posts (tweets) compared to the existing machine learning solutions?
    \item [\ding{113}] \textbf{RQ2:} How do different GPT products, each with their unique characteristics, perform in sentiment analysis tasks?
    \item [\ding{113}] \textbf{RQ3:} Can GPT models effectively address the complicated yet crucial linguistic sentiment-related nuances in the text, e.g., emotion, emojis, and mixed sentiment? 
\end{enumerate}
In our pursuit to answer these research questions, we conduct rigorous experiments using three distinct GPT models: instruct-based GPT, fine-tuned GPT, and embedding GPT. To benchmark the performance of these models, we choose the SemEval 2017 - Task 4~\cite{rosenthal2019semeval}, a reputable and widely used yardstick in sentiment analysis tasks. Further, we conduct a comparative assessment of our GPT models and previous models on this benchmark. Our study also explores linguistic nuances in state-of-the-art models and the best-performing GPT model, aiming to decipher the underlying reasoning behind their performance. In summary, our contributions are as follows:

\begin{itemize}
    \item [\ding{234}] To the best of our knowledge, we are the first to introduce an innovative application of GPT models within the context of sentiment analysis tasks.
    \item [\ding{234}] We facilitate a thorough comparison between GPT models and current methodologies employed in sentiment analysis.
    \item [\ding{234}] We compare and contrast various GPT models against one another.
    \item [\ding{234}] We perform an in-depth investigation of the competence of GPT in deciphering and processing linguistic nuances related to the sentiment of Tweets.
\end{itemize}

We anticipate that our findings will substantially enrich the extant comprehension of sentiment analysis, its inherent challenges, and plausible solutions. The implications of this research are anticipated to spur subsequent research endeavors and foster advancements within this domain.

The structure of this paper is as follows. Section~\ref{sec:related} surveys the trajectory of sentiment analysis, focusing on LLMs. Section~\ref{sec:methodology} introduces our methodological approach, delineating the potential of GPT models and our research questions. Section ~\ref{sec:experiments} describes the experimental design, including the SemEval 2017 - Task 4 dataset and our GPT model implementation. This section also presents our comparative analysis of GPT and traditional models. 
Finally, we conclude our study in Section~\ref{sec:conclusions}, summarizing our key findings, acknowledging the limitations of GPT, and contemplating potential directions for future research.

\section{Related Work}
\label{sec:related}
In this section, first, we review the sentiment analysis approaches and the progress made over the last fifteen years. Next, we review the emerging LLMs, such as GPT, and how they can play a role in text mining in general and sentiment analysis in particular. 
\subsection{Sentiment Analysis}
Sentiment Analysis, alternatively termed \textit{opinion mining}, is a crucial component of NLP. It concentrates on extracting and interpreting subjective data from a myriad of sources. The overarching objective is to comprehend the sentiment or stance of a speaker or writer regarding a specific topic or to discern the contextual polarity within a document~\cite{Syamala2019A}.

The evolution of sentiment analysis is marked by three pivotal phases: the Lexicon-based era, the Machine Learning (ML) era, and the present-day Transformer model era.

The nascent stages of sentiment analysis were dominated by lexicon-based techniques~\cite{hartmann2019comparing,netzer2019words}. These methods primarily involve calculating the number of positive and negative words within a text and then determining the overall sentiment based on these quantities. They use predefined lexicons, each word associated with a sentiment score~\cite{esuli2007sentiwordnet,pak2010twitter,thelwall2010sentiment,wilson2009recognizing}. Although these methods are straightforward and highly interpretable, they need improvement in comprehending the context and cope with complex linguistic structures. Further, they fail to adapt to the dynamic nature of language usage~\cite{Dean2018Sentiment}.

The advent of machine learning initiated a revolutionary phase in the field of sentiment analysis, shifting the emphasis from lexicon-based methodologies to data-driven techniques. The implementation of supervised learning methodologies, where models were trained on a labeled dataset and then utilized to predict sentiments of unseen data, rapidly gained traction~\cite{birjali2021comprehensive}. Notably, algorithms such as Naive Bayes, Support Vector Machines, and Random Forests have been shown to outperform lexicon-based techniques, especially in their adaptability to and learning capacity from the contextual nuances of text~\cite{mullen2004sentiment,turney2002thumbs,pang2004sentimental}. These early advancements signified the potential power of machine learning applications in sentiment analysis.

However, even with such progress, traditional machine learning techniques encountered hurdles in fully encapsulating the intricacies and nuances of human language~\cite{Guthier2017Language-independent}. This limitation led to further exploration and refinement of methods to capture the elusive complexity of sentiment within textual data. As the research evolved, machine learning classifiers were extensively applied to large-scale, real-world data sources like social media, demonstrating their versatility and wide applicability~\cite{kouloumpis2011twitter}. Such developments signify the continuous evolution of sentiment analysis methodologies, reflecting the significant role machine learning techniques continue to play in shaping this domain.

Presently, sentiment analysis is witnessing a paradigm shift with the emergence of Transformer-based models like BERT, GPT-3, and their variants~\cite{devlin2018bert,dale2021gpt}. These models leverage a deep learning framework that facilitates a superior understanding of context compared to their predecessors~\cite{pota2020effective,chiorrini2021emotion}. Their architecture, centered around a pre-training and fine-tuning paradigm, has demonstrated exceptional proficiency in sentiment analysis tasks, adeptly managing complexities such as negations, intensifiers, and implicit sentiments~\cite{praveen2023understanding,bello2023bert}. Furthermore, these models have proven their versatility across varied domains, underscoring their potential for future research in sentiment analysis~\cite{shaib2023summarizing,cohen2023enhancing}.

\subsection{Large Language Models}

Large language models have emerged as a cornerstone in NLP research because they can leverage extensive knowledge bases, efficient learning mechanisms, and a sophisticated understanding of intricate linguistic structures~\cite{Kim2019Probing}. These models employ computational models and statistical methods to capture and generate human-like text, thus enabling an exceptional comprehension of context and representation of complex ideas~\cite{radford2019language}. Their size, signified by the extensive number of parameters, contributes to the impressive ability to generate text that reflects human-like thought processes.
OpenAI, an esteemed research organization in artificial intelligence, has been pivotal in developing large language models. A notable contribution by OpenAI is GPT-3~\cite{brown2020language}, a state-of-the-art autoregressive language model. GPT-3, backed by a transformer architecture and a colossal parameter count of 175 billion, exhibits proficiency in generating text that mirrors human language remarkably~\cite{kolt2022predicting}. The model has been trained on a broad spectrum of internet text, resulting in a highly versatile tool that can produce contextually rich and coherent sentences, despite its performance being variable across tasks~\cite{chiu2021detecting}.

The scope of large language models' applications in processing textual data is broad and continually evolving. Ranging from machine translation~\cite{vaswani2017attention} and text summarization~\cite{liu2019text,kieuvongngam2020automatic} to dialogue systems~\cite{zhang2019dialogpt}, these models have demonstrated their versatility. The complex language understanding tasks like sentiment analysis~\cite{devlin2018bert} and emotion detection~\cite{mao2022biases} further showcase the applicability of these models, owing to their ability to generate human-like text.
Therefore, large language models are a vital area of exploration in NLP. Despite these models' remarkable capabilities, the diversity of their tasks calls for relentless exploration and enhancement of their capacities. Organizations like OpenAI, through contributions like GPT-3, significantly propel this research field, yet the quest for further advancement remains incessant.

Utilizing the GPT-3.5 model, our analytical framework is a groundbreaking approach that distinguishes itself from current methods in several fundamental ways. As the first deployment of GPT-3.5, our technology goes beyond the capabilities of earlier models and products, boasting superior comprehension and generation of text. Compared to other state-of-the-art transformer-based methods, such as RoBERTa, the GPT-3.5 method presents a unique advantage, particularly in social media content. The architecture and training of GPT-3.5 have equipped it with an advanced ability to comprehend linguistic nuances prevalent in social media content, a task that has traditionally proven challenging for other models. Its superior capacity to understand and generate contextually and semantically coherent responses in various conversational situations sets our method apart, positioning it as a revolutionary tool in language understanding and generation.

\section{Methodology}
\label{sec:methodology}
\begin{figure*}[htbp]
\centering
\includegraphics[width=1.0\textwidth]{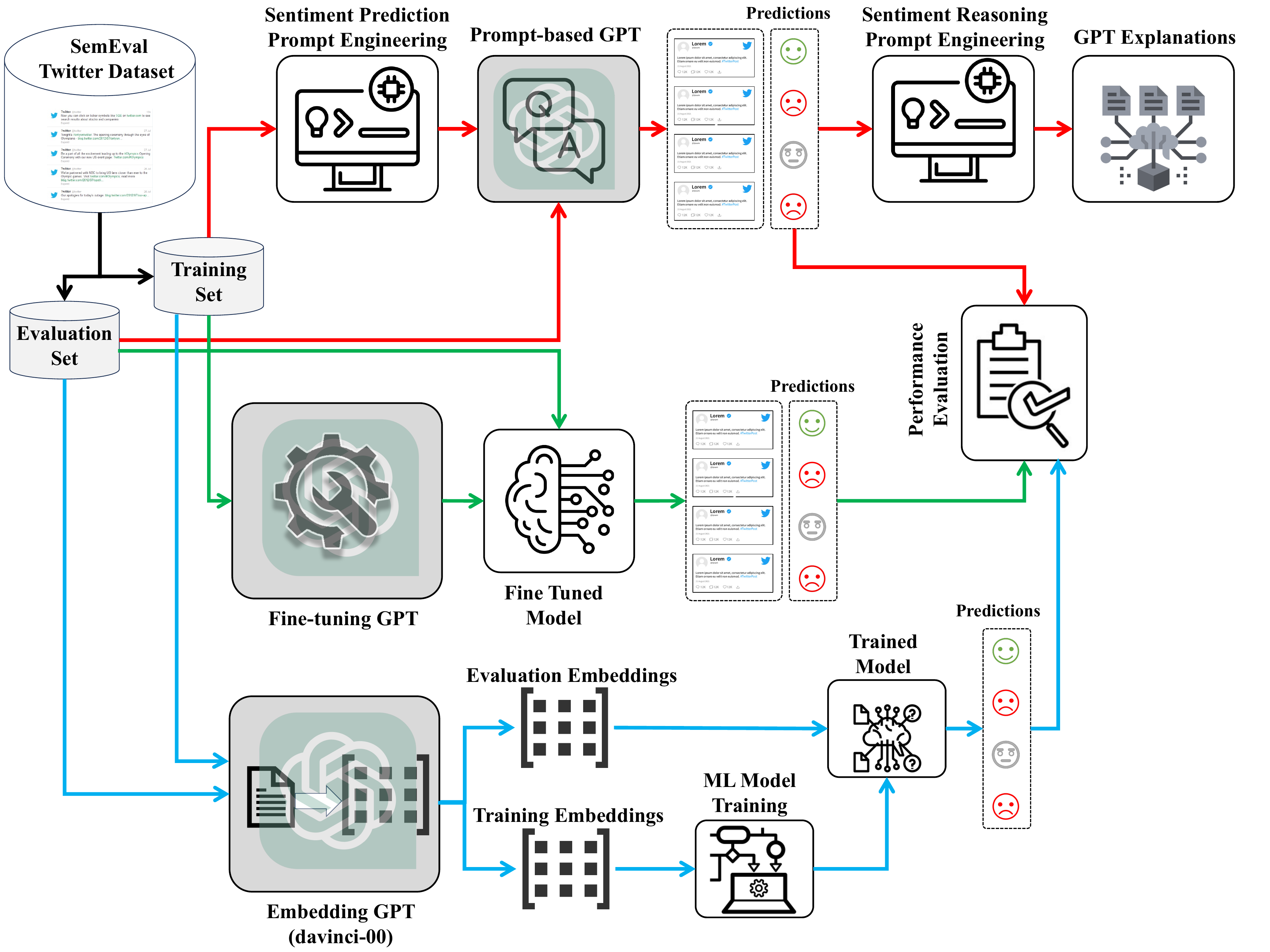}
\caption{An overall illustration of \texttt{SentimentGPT} framework}
\label{fig:framework}
\end{figure*}

Figure~\ref{fig:framework} illustrates a comprehensive depiction of the proposed framework, \texttt{SentimentGPT}, which leverages GPT for sentiment analysis. This system encompasses three primary components.

The initial component incorporates a prompt-based GPT approach for sentiment analysis. As shown in Figure~\ref{fig:framework}, we initiate the process with prompt engineering, crafting an effective prompt that enables GPT to perform sentiment analysis. Following this, GPT's predictions are compiled, and their performance is evaluated utilizing various metrics, such as the F1 score. To address \textbf{RQ3}, an additional prompt is engineered to extract explanations for the sentiment-based decisions that the prompt-based GPT has generated. The second component pivots to another OpenAI product, the fine-tuned GPT model. This model is trained on the training set and its associated ground truth labels, resulting in a fine-tuned output model. Predictions from this model are subsequently collected, and their performance is assessed. The final component harnesses the text embedding capabilities of GPT, extracting text embeddings from both the training and evaluation datasets. Following this, traditional machine learning models (e.g., Random Forest) are trained on the embeddings from the training set. Ultimately, these trained machine learning models are applied to the evaluation set embeddings, from which predictions are produced, and their performance is evaluated. Each of these components will be detailed further in the subsequent sections.

\subsection{Prompt-based Sentiment Prediction}

\subsubsection{Background}
Chat completion, also known as chatbot conversation modeling or dialogue generation, is a natural language processing (NLP) task involving the production of human-like responses in conversational contexts \cite{OpenAI_prompt}. Chat completion aims to create AI models capable of comprehending and generating appropriate responses in real-time conversations. To learn patterns, context, and the nuances of human conversation, these models utilize large language models trained on vast amounts of text data. The chat completion procedure typically involves using user input, such as a message or question, as a \textit{prompt} for the model. The model then responds based on its comprehension of the input and the conversational context. Typically, this response is generated by predicting the most probable sequence of words that would follow the given prompt, considering both grammar and semantic coherence.

As a chat completion model, GPT-3.5 models can understand and generate natural language or code. Their most capable and cost-effective model in the GPT-3.5 family is \textit{GPT-3.5-turbo}, which has been optimized for chat but works well for traditional completions tasks as well~\cite{OpenAI}. Although our original plan included a deep dive into GPT4, it is currently on a subscription model with a waitlist, limiting its accessibility. As a result, given its availability and relevance to our investigation, our analysis will primarily focus on GPT-3.5 Turbo.

\subsubsection{Sentiment Prediction Prompt Engineering}
We used an intricately crafted prompt proper for the GPT model, created using prompt engineering principles. This critical technique entails creating and refining input prompts to guide the responses of language models such as GPT-3.5 Turbo. It is similar to tinkering with a question until it yields the desired answer. For example, in sentiment analysis, prompts can be designed to steer the model toward a sentiment-oriented response rather than a purely factual one.

The prompt engineering journey begins with a simple question or statement that broadly encapsulates the desired response's direction. The model's response to this prompt is then scrutinized, and the prompt is tweaked and improved iteratively to produce better results. It essentially follows a cycle of refinement until the generated outputs match the task requirements. Nonetheless, successful prompt engineering follows a few guiding principles. The prompt should be clear and specific to reduce the likelihood of the model producing unrelated or overly generalized responses. Setting the context appropriately within the prompt can also guide the model toward the desired output style. It is also critical to balance the prompt's length, ensuring enough context is provided without prompting the model to generate overly long responses~\cite{liu2022design,white2023prompt}.
Include examples of desired responses within the prompt, especially in few-shot learning scenarios. This can direct the model to the desired output format. Finally, knowing the model's limitations, biases, and training data can help create more effective prompts.

In this study, we used the following carefully crafted prompt for the sentiment analysis task:

\begin{mdframed}[linewidth=1pt,linecolor=black,backgroundcolor=DarkOrchid!10, roundcorner=10pt]
\textit{``As a social scientist, Your task is to analyze the sentiment of a series of user tweets extracted from Twitter. Please assign a sentiment score from 0 to 2 for each tweet, where 0 signifies negative sentiment, 1 indicates neutral sentiment, and 2 corresponds to positive sentiment. In situations where the sentiment is difficult to definitively classify, please provide your best estimation of the sentiment score.''}
\end{mdframed} 

Utilizing the described prompt, we designed our training and test data sets to correspond to the required chat completion style.

\subsubsection{Performance Evaluation}
\label{subsub:performance}
Upon obtaining the model's predictions, we evaluated its performance. This evaluation includes several metrics such as accuracy, recall, and F1-score. It is important to note that even though our prompt specifically requested three classifications (positive, negative, and neutral), the GPT model retains the ability to return a ``mixed" sentiment if it determines that none of the predefined classes accurately represent the sentiment. We will discuss this more in the next section.

\subsubsection{Linguistic Nuances}

Investigating linguistic sentiment-related nuances, such as emotion, emojis, and mixed sentiment, is critical because these nuances contribute significantly to the overall sentiment expressed in a text. Such elements often carry additional layers of meaning beyond the plain text and can shift the sentiment context dramatically. In an era of digital communication, where text is frequently augmented with emojis and mixed sentiments are common, an effective sentiment analysis model must be capable of decoding these complexities. Hence, as mentioned before, \textbf{RQ3} is concerned with investigating whether GPT can detect these complex human language cues.  We hypothesize that GPT models, with their sophisticated language understanding capabilities, offer a promising avenue for such nuanced sentiment analysis. They are built upon a transformer architecture that allows them to better understand the context of words and sentences. Hence, these models could potentially detect and interpret these subtle sentiment-related cues more accurately than traditional models, thus providing a more precise sentiment classification. This capability is precious in various applications, such as social listening, customer feedback analysis, or trend prediction, where understanding subtle sentiment cues could lead to more accurate results and insights.

We reviewed academic papers on the challenges of sentiment analysis in traditional models, such as Hussein et al. (2018)~\cite{hussein2018survey}, Birjali et al. (2021)~\cite{birjali2021comprehensive}, and Wankhade et al. (2022)~\cite{wankhade2022survey}. Based on our research, we identified and classified 7 types of linguistic nuances.

\begin{itemize}
    
    \item [\ding{113}] \textbf{Emoji:} Traditionally, sentiment analysis has often relied on a simplistic theory suggesting that sentiment can be inferred solely based on the emojis used\cite{hussein2018survey}. However, this approach can be misleading and confusing at times. People's use of emojis may not always align with the sentiment expressed in their accompanying text. Emojis can be subjective and open to interpretation, making it challenging to rely solely on them for accurate sentiment analysis. To achieve more reliable results, it is necessary to consider the context, linguistic cues, and textual content in conjunction with emojis to gain a comprehensive understanding of sentiment in textual data.
        
    \item [\ding{113}] \textbf{Slang:} Numerous slang phrases, despite initially appearing neutral or even negative, carry positive connotations once their meanings are understood. Examples of such include expressions like ``the bee's knees," ``Cat's Pajamas," ``Dog's bollocks," ``Duck's nuts," and ``Real McCoy." Conversely, some phrases may seem innocuous but bear negative implications, such as ``Backhanded compliment," ``Throw someone under the bus," and ``A wolf in sheep's clothing." 
    
    \item [\ding{113}]  \textbf{Hashtag:} Hashtags often serve as valuable indicators to detect irony or sarcasm, elements notoriously challenging to discern in sentiment analysis. For instance, a seemingly positive statement might be accompanied by a hashtag such as \#sarcasm or \#not, implying the actual sentiment is indeed the reverse. While hashtags are primarily straightforward to decode and can be pretty enlightening, they might also introduce irrelevant data or noise. Alternatively, they could be employed unconventionally to draw attention rather than accurately portray sentiment. 
    
    \item [\ding{113}] \textbf{Negation \& Sarcasm:} One of the significant hurdles in sentiment analysis is handling negation and its interplay with sarcasm or irony\cite{Jia2009The}. Negation involves using certain words or phrases to counter an otherwise positive statement, effectively reversing the sentiment's polarity. For example, ``I am not happy today, but I was yesterday," the word ``not" negates the phrase ``happy today" but does not apply to ``I was yesterday." Determining the extent to which the negation applies within a sentence can be complex. In sarcasm or irony, negation can take a more nuanced role. For example, consider the sentence, ``Oh great, I just love working weekends." Here, despite using positive words like ``love" and ``great," the sentiment is genuinely negative. This usage underscores sarcasm, where the expressed sentiment contradicts the literal words. Recognizing and interpreting this contradiction poses a substantial challenge for machines, as they must learn to discern the implied tone and context that humans intuitively understand. Sarcasm often relies on shared knowledge and situational awareness\cite{Bharti2019Sarcasm}, making it an intricate facet of sentiment analysis, especially when it intersects with negation.

    \item [\ding{113}] \textbf{Mixed Sentiment:} Mixed sentiments present a complex difficulty in sentiment analysis as they embody both positive and negative sentiments within a single sentence or paragraph, making them difficult to classify. The level of granularity in sentiment analysis typically depends on the specific needs of the task at hand. Classifying a whole statement or paragraph as solely positive or negative becomes intricate when dealing with mixed sentiments. For instance, consider the sentence: ``I really love the camera on this new phone, but the battery life is awful." This sentence encapsulates positive sentiments toward the phone's camera and negative sentiments toward its battery life, which could confound an attempt to assign a unified sentiment label.

    \item [\ding{113}] \textbf{Cultural Context:} Cultural context forms a significant component of sentiment analysis as words and phrases can elicit varied sentiments across different cultures. The connotation and sentiment attached to a specific term might vary significantly from one culture to another, thereby complicating sentiment analysis. Cultural nuances, local dialects, colloquialisms, and region-specific slang can all significantly influence the interpretation of sentiment. Moreover, the task becomes even more complex when language translation is involved. When statements are translated from one language to another, there's a risk of losing or altering the original sentiment due to linguistic differences, the non-existence of certain words or phrases in the target language, or different cultural perceptions of the same concept. For instance, a complimentary phrase in one culture might be seen as an insult in another. Similarly, the tone of certain words or expressions, mainly when translated, can vary significantly, leading to changes in the perceived sentiment.

    \item [\ding{113}] \textbf{Modern Abbreviations:} Modern abbreviations and acronyms often present challenges in text analysis. However, advanced language models, such as GPT, demonstrate a higher degree of adaptability due to their training on extensive text corpora, enabling them to recognize and understand a wide array of new-age abbreviations. Table~\ref{tab:abbreviations} shows some notable examples. Incorporating an understanding of these abbreviations into sentiment analysis can significantly improve the accuracy and relevance of the results.
\end{itemize}

\begin{table}[htbp]\footnotesize
   \setlength{\extrarowheight}{1.5pt}
\setlength\tabcolsep{3pt} 
\caption{List of abbreviations and their meanings}
\centering
\begin{tabular}{|c|c||c|c|}
\hline
\textbf{Abbr} & \textbf{Meaning} & \textbf{Abbr} & \textbf{Meaning} \\ \hline
TBH & To Be Honest & ICYMI & In Case You Missed It \\ \hline
FOMO & Fear Of Missing Out & YOLO & You Only Live Once \\ \hline
SMA & Social Media Addict & OMW & On My Way \\ \hline
TIL & Today I Learned & AMA & Ask Me Anything \\ \hline
ELI5 & Explain Like I'm 5 & DM & Direct Message \\ \hline
SMH & Shaking My Head & IRL & In Real Life \\ \hline
BAE & Before Anyone Else & OOMF & \makecell{One Of My\\ Friends/Followers} \\ \hline
NSFW & Not Safe For Work & TL;DR & Too Long; Didn't Read \\ \hline
WYD & What You Doing & FTW & For The Win \\ \hline
NFT & Non-Fungible Token & RN & Right Now \\ \hline
OOTD & Outfit Of The Day &NGL & Not Gonna Lie \\ \hline

\end{tabular}
\label{tab:abbreviations}
\end{table}

\subsubsection{Selecting Tweets with Linguistic Nuances}
To comprehend the GPT model's performance and its interpretability, we identified examples from each category. For instance, to uncover challenges in the emoji groups, we sifted through the dataset, explicitly focusing on entries with emojis. Similarly, we employed a slang dictionary to filter the relevant data for slang terms. Hashtag-related issues were addressed by filtering posts using the `\#' symbol. For mixed-language entries, we singled out data that contained contrasting link words, and abbreviations were targeted by filtering based on abbreviated terms. However, identifying sarcasm and cultural references necessitated a more comprehensive approach, requiring us to review the evaluation set manually.

\subsubsection{Sentiment Reasoning Prompt Engineering}
\label{subsub:reasoning}
Throughout our analysis, we placed GPT-3.5 Turbo in contrast with other state-of-the-art models to highlight its unique advancements in sentiment analysis, and based on the following prompt, we asked for the reasoning behind the classification:

\begin{mdframed}[linewidth=1pt,linecolor=black,backgroundcolor=DarkOrchid!10, roundcorner=10pt]
  \textit{ ``Consider the following statement: `statement'. Please conduct a sentiment analysis on this statement. Could you identify the sentiment expressed in this statement and provide a detailed explanation of the rationale behind your assessment."}
\end{mdframed}

For every linguistic nuance we encountered, we provided sample comparisons and the reasoning behind them. Given that these linguistic challenges were resolved, it is plausible to anticipate that they contributed to enhanced accuracy. Additionally, this process gave us some insight into the model's interpretability.

\subsection{Fine tuning}
\subsubsection{Background}
Fine-tuning is a crucial process within the GPT API that involves further training a pre-trained language model to specialize in specific domains or tasks by utilizing domain-specific or task-specific datasets. The GPT models were initially trained on a vast corpus of internet text to grasp grammar, language patterns, and general knowledge. However, fine-tuning allows these models to refine their capabilities by exposing them to more targeted training data.

\subsubsection{Fine-tuning GPT Procedure}
The process of fine-tuning starts with taking the pre-trained GPT model and training it on a smaller dataset that is specifically relevant to the desired task or domain, such as legal texts or medical literature. We evaluated three models for our task: Ada, Babbage, and Curie. The pricing structure, features, and capabilities of these models differ.

\textbf{Ada.}
Ada is the largest and most advanced model among the three, offering high capacity and exceptional performance across various language tasks. Its many parameters enable it to handle complex and nuanced language processing tasks effectively. Ada's capabilities make it suitable for demanding applications that require state-of-the-art performance. However, its larger size may come with increased cost and computational requirements.

\textbf{Babbage.}
 Babbage strikes a balance between performance and cost. While it has slightly reduced capacity compared to Ada, it still offers competitive performance in various language tasks. Babbage is a cost-effective option for applications that require reliable language processing capabilities without the need for the highest level of performance. It provides a good balance between cost-efficiency and computational requirements.
 
\textbf{Curie.}
Curie is the smallest model among the three but still offers reliable language processing capabilities. Compared to Ada and Babbage, it may have limitations in handling more complex or nuanced language tasks. However, Curie's smaller size makes it a cost-effective choice, particularly for applications with resource constraints. It provides a reasonable level of performance at a reduced cost and computational requirement compared to larger models.

\subsubsection{Performance Evaluation}
We adapted our training and evaluation sets to conform to the format required by the fine-tuned models. Subsequently, these pre-trained models were fine-tuned using the adapted training set. Their performance was then assessed using the adjusted evaluation set. This assessment adhered to the methods outlined in Section~\ref{subsub:performance} but with one key distinction: the models were instructed to provide predictions based on only the three predefined classes - positive, negative, and neutral - and to refrain from predicting the 'mixed' sentiment.

\subsection{Embedding GPT }
\subsubsection{Background}
The term ``embedding" refers to transforming words or other textual inputs into numerical representations, also referred to as ``embedding". GPT models, which are based on the architecture of the transformer\cite{gpt_embedding}, can extract a wealth of semantic and contextual information from text.
The GPT model creates an embedding for each token (word or subword, depending on the tokenization strategy) that captures the semantic and syntactic essence of the token. Tokenization is the process by which each token is initially assigned a distinct numeric identifier. Using a trainable parameter called an embedding layer, these identifiers are converted into embeddings-- dense vectors of a fixed size. Embeddings in the GPT model are attractive because of their capacity to record detailed linguistic data. For instance, the model can generalize well on language tasks because words with similar meanings have embeddings close to each other in the vector space.

To consider the sequence in which tokens occur, the GPT model additionally employs positional embeddings\cite{gpt_embedding}. Positional embeddings are essential for providing a sense of sequence or temporal information because GPT's building blocks, transformers, do not understand the order of tokens by default. The GPT model's input text representation consists of a combination of token and positional embeddings. Once these embeddings are generated, they are fed into a series of transformer blocks, where they are further refined and manipulated better to capture the intricate interplay between tokens and their environments.

\subsubsection{Embedding Acquisition}
For getting the embeddings of train and test, we used the model \textit{text-embedding-ada-002}, which is the recommended model for embedding in most tasks and is cheaper than other models.
To utilize GPT embeddings for sentiment classification, the following steps were taken:
\begin{enumerate}

  \item \textbf{Preprocessing:} Preprocess the textual data by removing noise, normalizing text, and handling punctuation, stop words, and special characters. Tokenize the text into individual words or subwords.

  \item \textbf{Obtaining GPT Embeddings:} Utilize a pre-trained GPT model to generate embeddings for each word in the text. Pass the tokenized words through the GPT model, and extract the corresponding embeddings from a specific layer or layers of the model.

  \item \textbf{Dimensionality Reduction:} To effectively utilize the embeddings within our machine learning models, it was necessary to reduce the dimensionality of the embeddings. To accomplish this, we employed the principal component analysis (PCA) technique, which allowed us to mitigate the risk of overfitting the model. We reduced the dimension of embedding from 400 to 150. 
\end{enumerate}
\subsubsection{ML Model Training}.
Using these embeddings as inputs, we trained machine learning models for sentiment classification tasks by employing the embeddings of the training set-- See Figure~\ref{fig:framework}. We trained  XGboost and Random Forest (RF) models for classification.
\subsubsection{Performance Evaluation}
 Then, utilizing the evaluation datasets, we assessed the performance of these models (using accuracy, recall, and F1-score) and analyzed the results in depth to determine the efficacy of GPT embeddings in performing sentiment classification. 
\section{Experiments}
\label{sec:experiments}
In this section, we delineate the experiments. We initially provide details of the dataset in Section~\ref{subsec:dataset}. Subsequently, we elaborate on the specific experimental configurations in Section~\ref{subsec:settings}. Ultimately, in Section~\ref{subsec:results}, we present the outcomes of our experiments and engage in a comprehensive discussion to address the research queries that were previously posited.

\subsection{DataSet}
\label{subsec:dataset}
In this study, we have utilized the SemEval-2017 Task 4 dataset, a collection of tweets on specific subjects, primarily in English. SemEval, originally SensEval, is an international workshop series that promotes semantic evaluation. Its primary task, sentiment analysis on Twitter, was introduced in 2013 and mainly focused on discerning sentiment polarity in a text, regardless of whether it is general or topic-specific. The SemEval-2017 Task 4 dataset improves upon the previous years' model by including Arabic language content to enhance multilingual and cross-lingual sentiment analysis. However, our study limits its scope to English language content and does not delve into Arabic. However, it should be noted that the SemEval-2017 Task 4 dataset is imbalanced regarding the distribution of tweets' sentiment polarity. In particular, when analyzed on a 3-point scale (our chosen scale), there are more tweets classified as ``Positive" and ``Neutral" than ``Negative"-- See Table~\ref{tab:dataset stat}. This imbalance poses a unique challenge for current machine learning sentiment analysis models. Moreover, SemEval-2017 Task 4 dataset includes annotations for detecting sentiment with different granularity levels, including 2-point, 3-point, and 5-point scales, presented across five subtasks (A to E). However, this study only focuses on the 3-point scale, not the other scales or subtasks.
\begin{table}[H]
    \centering
     \caption{Basic dataset statistics}
    \begin{tabular}{|c|c|c|c|c|}
    \hline
       \textbf{Split}& \textbf{Positive} & \textbf{Neutral} & \textbf{Negative}& \textbf{Total} \\
       \hline
       Train  & 19,902 &22,591&7,840 &50,333 \\
       Test  &2375 &5,937&3,972&12,284\\
       \hline
    \end{tabular}
    \label{tab:dataset stat}
\end{table}

\subsection{Experimental Settings}
\label{subsec:settings}
We employed Python as our primary programming language for our GPT-based models, including GPT 3.5 Turbo, Ada, Babbage, Curie, and Davinci. We used OpenAI API and GPT's command line interface (CLI) for our experimentation and model training. The models' configuration was set for high precision, with a temperature setting 0. This setting ensures that the models generate outputs with the highest predicted probabilities. It encourages the models to generate more focused, deterministic outputs. To ensure the robustness and reliability of the results, we used a \textit{best of 3} setting for the GPT models. This means the models were run three times for every prediction, and the most frequently predicted output was selected as the final prediction.  

We used PyCaret, an open-source, low-code machine learning library in Python, to develop The GPT Embedding + Xgboost and GPT Embedding + Random Forest models. PyCaret \cite{gain2021low} simplifies the process of creating and deploying machine learning models and is ideal for this experiment due to its efficiency and ease of use. 
The detailed hyperparameters for the XGboost and Random Forest models were optimized using PyCaret's automated machine learning capabilities, which simplified the process of hyperparameter tuning and model selection. This setup allows the models to take advantage of the powerful representational ability of GPT, coupled with the classification performance of XGboost and Random Forest algorithms.

\subsection{Experimental Results}
\label{subsec:results}
In this part, we present and analyze the results obtained from the experiments. We carefully evaluate the performance of each model and select the best-performing
one for further comparison.  We consider different metrics, including accuracy, precision, recall, and F1 score, to assess the performance of GPT variants and prior models. Additionally, we provide explanations and insights into the reasons behind the superiority of the selected model, highlighting its noteworthy features and performance compared to other models. Through this evaluation process, we strive to answer \textbf{RQ1}, \textbf{RQ2}, \textbf{RQ3} and obtain a robust and reliable sentiment classification model that can accurately predict sentiment in real-world scenarios. 

\subsubsection{GPT Variants and Comparison with Baselines}

As the baseline methods, we considered the following, all trained and evaluated on the SemEval-2017 dataset.

\begin{itemize}
    \item[\ding{113}]  Baziotis et al.~\cite{baziotis-etal-2017-datastories-semeval}:
    This paper utilized the MSA model, a two-layer BiLSTM (Bidirectional Long Short-term Memory Networks) network with an attention mechanism. Pre-trained word embeddings embed Twitter messages into a low-dimensional vector space. The BiLSTM layers capture and summarize sentence information from both directions, and an attention layer assigns word importance based on its contribution to the message's sentiment. The weighted sum emphasizes the most sentimental words in the message. 
    \item [\ding{113}] Deshmane et al~\cite{deshmane-friedrichs-2017-tsa}: Their paper model used three sentiment-analysis-focused deep learning architectures.  The first component is a CNN (Convolutional Neural Network), which classifies text well. Gated recurrent neural networks (GRNNs) can detect temporal dependencies in sequential data like text in the second subsystem. Finally, the third system uniquely integrates opinion lexicons, databases of sentiment-expressing words, directly into another CNN architecture. Combining these three systems should improve sentiment analysis.

    \item [\ding{113}] Cliche et al~\cite{cliche2017bb_twtr}:
    In this paper, the authors made a Twitter sentiment classifier using Convolutional Neural Networks (CNNs) and Long Short Term Memory (LSTMs) networks. To boost performances, they ensembled several CNNs and LSTMs together. 
    
    \item [\ding{113}] Hao et al.~\cite{hao-etal-2017-xjsa}: Word2Vec was used to obtain an embedding of the tweets, and a 3 layer CNN, SSWE(sentiment-specific word embedding)~\cite{tang2014learning}, was then used to analyze the data.
    \item [\ding{113}] Gonzalez et al.~\cite{gonzalez-etal-2017-elirf}: Their approach was based on the use of convolutional and recurrent neural networks and the combination of general and specific word embeddings with polarity lexicons.
    \item [\ding{113}]Nguyen et al.~\cite{nguyen2020bertweet} (RoBERTa): BERTweet, having the same architecture as BERT-base \cite{devlin2018bert}, was trained using the RoBERTa pre-training procedure \cite{liu2019text}. 
\end{itemize}

\begin{table*}[htbp]\normalsize
    \centering
    \caption{Sentiment Analysis Performance across GPT Variants and Previous Models}
    \begin{tabular}{c|c|c|c|c}
        &\textbf{Model} & \textbf{Accuracy} & \textbf{Recall} & \textbf{F1-score} \\
        \hline
       \multirow{ 7}{*}{\rotatebox[origin=c]{-90}{GPT Variants} }& GPT 3.5 Turbo & \textbf{0.9732} & 0.9198 & 0.9426 \\
        & Ada & 0.9528 & 0.9306 & \textbf{0.9473} \\
        & Babbage & 0.9594 & \textbf{0.9653} & 0.9084 \\
        &Curie & 0.9316 & 0.8927 & 0.9149 \\
        & Davinci & 0.9243 & 0.8531 & 0.8809 \\
        & Embedding + Xgboost & 0.8736 & 0.8405 & 0.8631 \\
        & Embedding + RF & 0.8348 & 0.8192 & 0.8729 \\
            \Xhline{3\arrayrulewidth}
        \multirow{ 7}{*}{\rotatebox[origin=c]{-90}{\makecell{Previous \\ Methods}}} & Baziotis et al.~\cite{baziotis-etal-2017-datastories-semeval} & 0.651 & 0.681 & 0.677 \\
        &Deshmane et al.~\cite{deshmane-friedrichs-2017-tsa} & 0.616 & 0.643 & 0.620 \\
        &Cliche et al.\cite{cliche2017bb_twtr} & 0.658 & 0.681 & 0.685 \\
        &Hao et al.~\cite{hao-etal-2017-xjsa} & 0.575 & 0.556 & 0.519 \\
        &Gonzalez et al.~\cite{gonzalez-etal-2017-elirf} & 0.599 & 0.632 & 0.619 \\
        &Nguyen et al.~\cite{nguyen2020bertweet} (RoBERTa) & \textbf{0.720} & \textbf{0.732} & \textbf{0.725} \\
        
    \end{tabular}
    
    \label{tab:combined_model_comparison}
\end{table*}

Aiming at answering \textbf{RQ1} and \textbf{RQ2}, we make the following observations based on the results presented in Table~\ref{tab:combined_model_comparison}.

\begin{itemize}
    \item [\ding{237}] GPT-related models significantly outperform existing machine learning solutions in sentiment analysis tasks for social media posts (tweets). This is evident across all evaluated metrics - Accuracy, Recall, and F1-score.
    \item [\ding{237}] Superior performance of the GPT models can be attributed to their advanced transformer architectures, large-scale training, and capability to understand the context in text data.
    \item [\ding{237}] The Recall and F1-score metrics follow a similar trend, with GPT variants consistently outperforming the previous methods.
    \item [\ding{237}] Among the different GPT products, GPT 3.5 Turbo achieves the highest Accuracy, while Babbage obtains the highest Recall of 0.9653, and Ada scores the highest F1-score of 0.9473. This suggests that different GPT variants may excel in different aspects of sentiment analysis tasks.
    \item [\ding{237}] High accuracy, as demonstrated by GPT 3.5 Turbo, indicates a lower error rate, suggesting superior model performance.
    \item [\ding{237}] A high recall score, such as Babbage's, denotes the model's aptitude to correctly identify the majority of sentiments in the dataset, including positive, negative, and neutral sentiments.
    Item [\ding{237}] A high F1-score (harmonic mean of precision and recall), like the one exhibited by Ada, suggests it is the best model when the costs of false positives and false negatives are very high.
    \item [\ding{237}] GPT variants combined with machine learning models, such as Xgboost and Random Forest (RF), also show commendable performance but with slightly lower scores on the evaluated metrics compared to standalone GPT models.
\end{itemize}

\begin{figure}[htbp]
    \centering
    \includegraphics[width=\columnwidth]{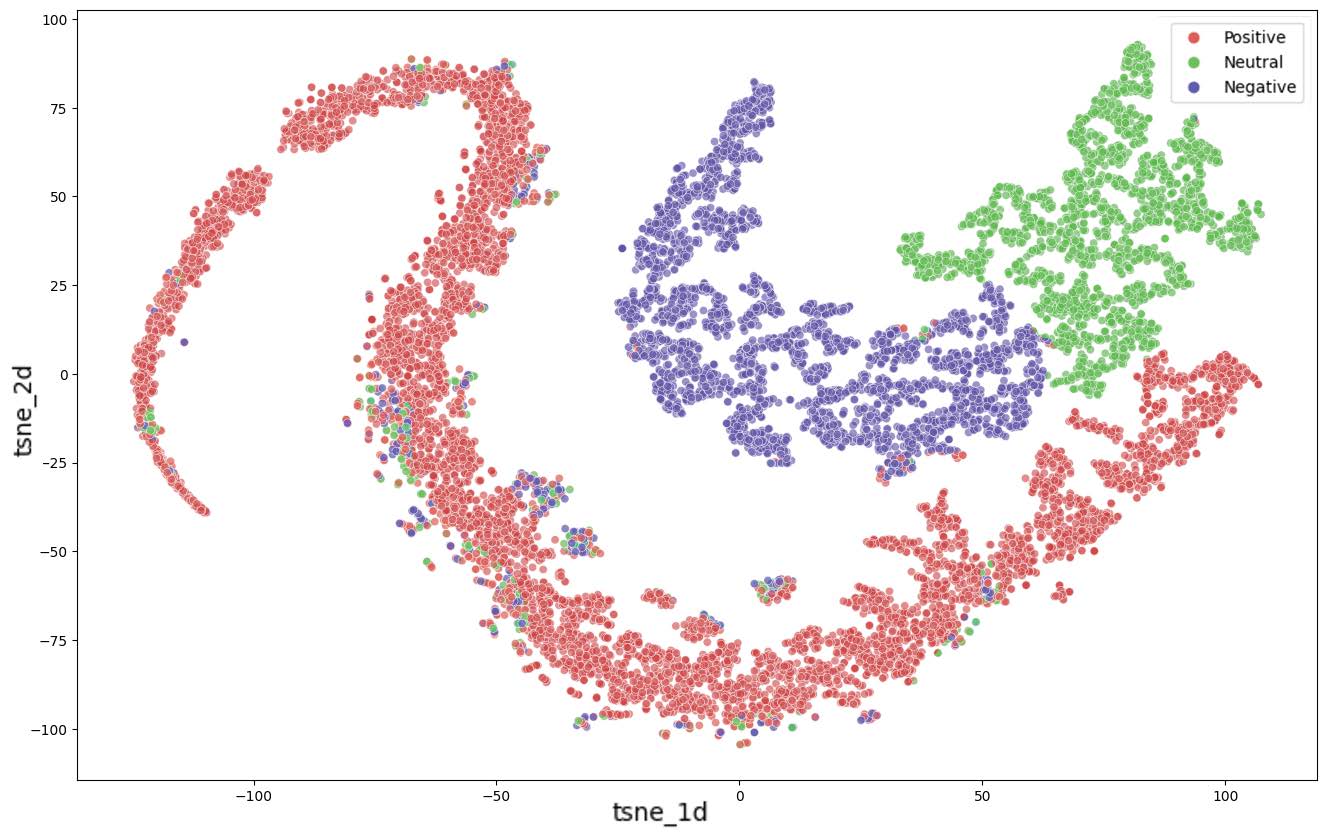}
    \caption{TSNE visualization of GPT embeddings}
    \label{fig:TSNE}
\end{figure}

Figure~\ref{fig:TSNE} demonstrates the visualization of GPT embeddings using the t-SNE technique~\cite{hinton2002stochastic}. This figure shows that the embeddings of different classes are distinctly separated into different regions (except in some cases). This again exhibits the power of GPT to learn salient representations of text, especially from tweets where the text is noisy and short.     

\subsubsection{GPT reasoning and linguistic nuances in Twitter}

    \begin{table*}[htbp]\footnotesize
       \setlength{\extrarowheight}{1.5pt}
\setlength\tabcolsep{2pt} 
    \centering
    \caption{GPT-3.5 vs. RoBERTa models in sentiment analysis of tweets containing  linguistic nuances (N: Neutral, M: Mixed)}
    \begin{tabularx}{\textwidth}{|c|p{6cm}|p{0.25cm}|p{0.25cm}|p{0.25cm}|X|}
    \hline
    \rotatebox[origin=c]{-90}{\textbf{Category}}&\textbf{Tweet} & \rotatebox[origin=c]{-90}{\textbf{GPT-3.5}}  & 
    \rotatebox[origin=c]{-90}{\textbf{RoBERTa}} &  \rotatebox[origin=c]{-90}{\textbf{True Label}}&\adjustbox{valign=m}{\textbf{GPT Reasoning}}\\
    \hline
      \multirow{4}{*}{\rotatebox[origin=c]{-90}{Emoji}} &I would love to die now more than anything else. \adjustbox{valign=b}{\includegraphics[width=0.025\textwidth]{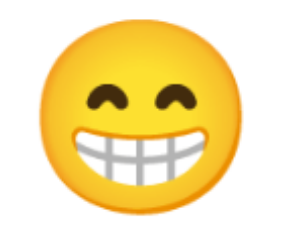}} & -  &  +  &-&The negative sentiment in the statement stems from the expression of a desire for death, despite the presence of a positive emoticon at the end.\\
      \cline{2-6}
      &My best friend is moving to another city \adjustbox{valign=b}{\includegraphics[width=0.02\textwidth]{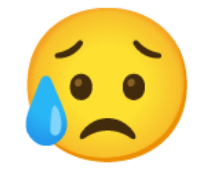}}
 & -& N &-& It expresses sadness or a sense of loss, which are considered negative emotions. \\
            \cline{2-6}
            &I tripped and fell flat on my face in front of everyone \adjustbox{valign=b}{\includegraphics[width=0.02\textwidth]{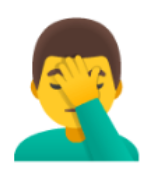}}& - & +&-& The text recounts an embarrassing incident of tripping and falling in public, which is generally regarded as a negative experience. \\
            \cline{2-6}

      &I'm staying up all night to finish this report. \adjustbox{valign=b}{\includegraphics[width=0.065\textwidth]{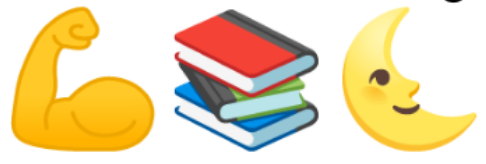}} & + & - &+& It communicates determination, resilience, and commitment to completing a task, which are generally seen as positive attributes. \\
     
    \Xhline{3\arrayrulewidth}
    \multirow{4}{*}{\rotatebox[origin=c]{-90}{Slang}}&She called the new product the dog's bollocks. &+&-&+& The sentiment of the sentence is positive because ``the dog's bollocks" is British slang for something outstanding or excellent.\\ 
    \cline{2-6}
     &@user You're worse than the ``alt-rightists", they are bold enough to display their bigotry for the world to see, but not I'll 'Billie'.&-& N & - &It includes derogatory language (``worse," ``bigotry") and expresses disdain or criticism towards righties political group. \\
        \cline{2-6}
    &@user leftists have cried wolf a few too many times, while Trump supporters r getting assaulted actually.  & -& N& - &``leftists have cried wolf," which implies that the speaker believes left-leaning individuals have made false claims or exaggerated issues.\\
        \cline{2-6}

    &Your dress looks pretty cheap but suits you well. Such backhanded compliments& -&+&-&The implication that the dress looks cheap is the insulting or negative aspect, even though it's followed by a statement that the dress suits the recipient well.\\
    \Xhline{3\arrayrulewidth}
    &@user ...and by ``genocidal antisemites" you mean Steve Bannon, right?\#NotMyPresident & -&N& -& the hashtag ``\#NotMyPresident" suggests a negative sentiment towards the person being referred to as president.\\
    \cline{2-6}
    \multirow{4}{*}{\rotatebox[origin=c]{-90}{Hashtag}} &Who needs sunshine when you can have a week full of rain? \#LiquidSunshine&-&N&-&It reflects disappointment or frustration with continual rain, despite the playful hashtag.\\
    \cline{2-6}
    &Isn't it great when your Wi-Fi decides to go on vacation in the middle of a workday? \#DigitalDetox&-&+&-& It conveys frustration over an unexpected and disruptive internet outage during a workday, despite the sarcastic tone.\\
    \cline{2-6}
    &A rainy day again, stuck indoors with a pile of books \#BookwormHeaven&+&-&+& Despite the potentially negative implications of rain and being stuck indoors, the phrase `\#BookwormHeaven' suggests the speaker enjoys reading and is happy with the situation.\\
    \Xhline{3\arrayrulewidth}
    \multirow{4}{*}{\rotatebox[origin=c]{-90}{Negation \& Sarcasm}}
   &Fake news, Election Day hacks, Wikileaks, Comey, Russia, Jill Stein, Gary Johnson, etc. are all to blame, but Clinton herself, isn,&-&N&-&The author is expressing dissatisfaction and disbelief about the perceived lack of blame placed on Clinton amidst a host of other blamed entities and events.\\
    \cline{2-6}
    &How many deaths will it take `till he knows that too many people have died? The answer, my friend, is blowin' in the wind...-Bob Dylan.&-&+&N&It highlights the idea of death and too many people dying in sarcastic way.\\
    \cline{2-6}
    & @user not only did he not say it. It was hard enough for american born people to avoid illegal voter suppression to vote &-&+&-&The speaker expresses a belief that the person mentioned did not make a certain statement.\\
    \cline{2-6}
    &I am not happy today, and I wasn't yesterday either& -&N& -&The speaker explicitly states their unhappiness, which is a negative emotion, across two days.\\
    \Xhline{3\arrayrulewidth}
    \multirow{4}{*}{\rotatebox[origin=c]{-90}{Mixed Sentiment}} &Spirit of my silence i can hear you. But im afraid to be near you. \#ThisIsUs& M &-&N&The sentiment in this text can be interpreted as mixed, with the user expressing both longing and fear towards an unidentified entity or situation, possibly tied to the show ``This Is Us".\\
    \cline{2-6}
   &1st job at 14, retired 65. In spite of short comings being female, managed white male supremacists \& became success.&M&+&N& While they mention success, the inclusion of the term ``shortcomings" implies that there were challenges along the way. \\
    \cline{2-6}
    &@user it's weird 'cause I usually like music in his vein, but it's every so shy of being memorable like a Bob Dylan/Lou Reed thing(?)&M &N&N& 
The tweet reflects appreciation for the artist's style, yet disappointment in its lack of memorability.\\
    \cline{2-6}
    &idk where to start after my intro. im writing from a pro-raising-minimum-wage pov but I'm stuck&M &-&N&The user expresses enthusiasm for a pro-raising-minimum-wage viewpoint (positive sentiment), but also experiences difficulty in progressing their writing (negative sentiment).
\\
    \Xhline{3\arrayrulewidth}
    \multirow{3}{*}{\rotatebox[origin=c]{-90}{Cultural }} &The performance was so sick.&+&-&+&`Sick' is often used to mean `outstanding' or `impressive', indicating approval or admiration.\\
    \cline{2-6}
    &@user The Leftists flock like sheep. Not an individual thinker amongst them.&-&N&-&The speaker uses derogatory language (``The Leftists flock like sheep") to describe individuals with left-leaning political beliefs implies a lack of independent thinking. \\
        \cline{2-6}
    &Just aced my test. \#dead&+&-&+&The sentiment is positive; ``aced my test" shows success,``\#dead" implies relief.\\
    \Xhline{3\arrayrulewidth}
     \multirow{3}{*}{\rotatebox[origin=c]{-90}{ Abbreviations}}&POC representation in films is just PC culture gone mad.&-&N&-&It dismisses importance of racial representation and diversity.\\
    \cline{2-6}
     &IMHO, this is one of the cheapest films of the decade.&-&N &-& ``IMHO" (In My Humble Opinion) implies personal viewpoint, ``cheapest" commonly denotes low quality, indicating dissatisfaction.\\
    \cline{2-6}
    & @user babe i thought you can do whatever you want... But, bae, I'm no longer want to be without you.&+&N&+&Speaker uses endearing terms like ``babe" and ``bae" to address the other person and expresses a desire to no longer be without them, indicating a positive.\\
    \hline
    \cline{1-6}
    \end{tabularx}
    \label{tab:challenge}
    \end{table*}

\textbf{RQ3} is concerned with crucial linguistic nuances in Tweets as to whether GPT can discern these cues while making sentiment predictions. We enumerated and explained seven linguistic cues in Section~\ref{subsub:reasoning}. Table~\ref{tab:challenge} shows the results of the experiments. Here, we showcase several tweets in each category, along with GPT-3.5's predictions (prompt-based GPT), RoBERTa's predictions,  true labels, and GPT reasonings. We make the following observations based on the results presented in this table. 

\begin{itemize}

\item [\ding{237}] \textbf{Emoji:} GPT-3.5 can interpret the context of the tweet instead of just focusing on the emojis used. For example, in the tweet ``I would love to die now more than anything else.\adjustbox{valign=m}{\includegraphics[width=0.025\textwidth]{Figures/Grin.png}}", the presence of the positive emoticon could be misleading, but GPT-3.5 correctly identified the negative sentiment of the text. In comparison, RoBERTa failed to identify the sentiment in all four cases correctly.

\item [\ding{237}] \textbf{Slang:} GPT-3.5 consistently identified the sentiment of the text correctly, even when slang expressions were used. In contrast, RoBERTa seemed to struggle, failing to identify the sentiment in half of the cases correctly. This indicates that GPT models are better equipped to understand and interpret slang expressions.

\item [\ding{237}] \textbf{Hashtag:} GPT-3.5 correctly interpreted the sarcasm implied by hashtags such as `\#NotMyPresident' in the context of the tweet. While RoBERTa missed the intended sentiment in all four cases, GPT-3.5 accurately recognized the sentiment in all cases, despite the potentially misleading hashtags.

\item [\ding{237}] \textbf{Negation \& Sarcasm:} In this category, it is noteworthy to highlight the performance of GPT-3.5 in identifying elements of sarcasm and negation present within the tweets, a task at which RoBERTa was unable to succeed. Interestingly, GPT-3.5 has demonstrated its superior reasoning capabilities in the context of the second tweet, where it successfully discerned the sarcastic—and thus, inherently negative—tone, even though the human-assigned ground truth label was neutral. This showcases GPT-3.5's remarkable ability to detect potentially mislabelled instances, underscoring its power and potential for nuanced understanding in natural language processing tasks.

\item [\ding{237}] \textbf{Mixed Sentiments:} GPT-3.5 has consistently demonstrated its capability to discern mixed sentiments across all cases, indicating its robust ability to comprehend and navigate the intricacies of complex emotional content. Conversely, RoBERTa faced difficulties accurately identifying the sentiment within these tweets, incorrectly interpreting the sentiment in each instance. Further highlighting the proficiency of GPT-3.5, the rationale it provided appeared logical and sensible, even when the human annotators had assigned a neutral label.

\item [\ding{237}] \textbf{Cultural Context:} GPT-3.5 has showcased its robust capacity to comprehend cultural contexts, as evidenced by its correct sentiment identification in phrases such as \textit{sick} or \textit{flock like sheep}. In contrast, RoBERTa's performance was marked by misinterpretations in all three instances, suggesting potential challenges in its ability to grasp cultural nuances.

\item [\ding{237}] \textbf{Modern Abbreviations:} GPT-3.5 consistently identified the sentiment accurately in all instances, contrary to RoBERTa, which persistently misinterpreted the sentiment. This implies that GPT-based models possess superior adaptability in comprehending modern abbreviations and the corresponding sentiment implications compared to their counterparts.

\end{itemize}

In summary, GPT-3.5 better understands the intricate aspects of linguistic sentiment-related nuances, such as emojis, slang, hashtags, sarcasm, mixed sentiments, cultural context, and modern abbreviations, compared to RoBERTa. It is better equipped to look beyond the surface-level sentiment and analyze the more profound, often complex sentiments that can be present in the text.

\section{Conclusion}
\label{sec:conclusions}

\subsection{Summary}
The advent of Generative Pretrained Transformer (GPT) models has immense benefits for sentiment analysis, which is essential for making sense of and taking action on massive amounts of user-generated data. This comprehensive analysis of GPT for sentiment analysis used the SemEval 2017 dataset to examine numerous methods. The study highlighted the benefits of GPT models. Three major research questions were posed. To answer these research questions, we employed a three-pronged strategy: prompt-based GPT, a fine-tuned GPT model, and GPT embedding classification. We demonstrated that state-of-the-art models, particularly GPT-3.5 Turbo, outperformed their fine-tuned, whole-dataset-trained counterparts by a wide margin. The performance of the GPT models was also compared to that of some baselines to demonstrate their superiority. This research also emphasized the remarkable effectiveness of GPT models in overcoming the standard linguistic nuances of sentiment analysis, such as contextualization and sarcasm detection. It became clear that these models' strengths lay in their ability to understand and analyze sentiment across a wide range of contexts, with a focus on their ability to deal with the ambiguities, negations, slang, and contemporary abbreviations that are so pervasive in modern social media language. These are issues that have long hampered progress in sentiment analysis.
\subsection{GPT Challenges}
\begin{itemize}

\item [\ding{113}] \textbf{Privacy:} Data privacy is a significant obstacle for GPT models. In order to make reliable forecasts and decisions, these models typically need access to extensive data sets, which may include sensitive information. This gives rise to worries about the confidentiality of data and the risk of its improper use or handling. Maintaining user privacy while reaping the benefits of GPT models is an ongoing challenge that calls for strong privacy protection mechanisms and well-defined policies for data management.

\item [\ding{113}] \textbf{Social Bias:} Since GPT models are trained using massive amounts of text data, any inherent biases in the training data can seep into the model's predictions. This can lead to unfair or biased results, which can feed into and strengthen preexisting prejudices and biases in society~\cite{liu2021authors}. It is crucial to ensure fairness, inclusivity, and unbiased decision-making by addressing and mitigating social bias in GPT models. Efforts are being made to incorporate diverse and representative datasets to counteract bias propagation, and methods are being developed to detect and mitigate bias during the training process\cite{social_bias}.

\item [\ding{113}] \textbf{Cost:} GPT models are computationally expensive because of the large amount of processing time, data storage, and electricity they require. Improving performance by scaling these models up can be expensive. Furthermore, additional computational resources and labeled data are required for fine-tuning GPT models on specific tasks. Finding a happy medium between model performance and the costs associated with training, deploying, and maintaining GPT models is a challenge researchers and organizations face.
\end{itemize}
The abovementioned challenges underscore the importance of careful planning, rigorous testing, and ongoing ethical considerations when working with advanced AI models such as GPT. As AI continues to evolve, we must address these challenges to ensure the development of equitable, accessible, and effective AI systems.

\subsection{Future Work}
While this study presents considerable advancements in applying GPT models for sentiment analysis, there remains an extensive scope for further exploration. Here, we outline several potential avenues for future research that could further our understanding and broaden the application of these models in the sentiment analysis domain.

\begin{itemize}
    \item [\ding{113}] \textbf{Contextual Sentiment Analysis:} Explore GPT models' performance on sentiment analysis in specific contexts such as crisis response, e-commerce reviews, or political discourse.
    
    \item [\ding{113}] \textbf{Adaptation to Different Languages:} Investigate the performance of GPT models on sentiment analysis in languages other than English, focusing on the unique challenges each language presents.

    \item [\ding{113}] \textbf{Emotion Recognition:} Extend the research to delve into emotion recognition, distinguishing finer shades of sentiment like joy, anger, sadness, etc.

    \item [\ding{113}] \textbf{Enhanced Sarcasm and Irony Detection:} Develop specialized modules or strategies to improve sarcasm and irony detection in sentiment analysis.
    
    \item [\ding{113}] \textbf{Fairness and Bias in Sentiment Analysis:} Explore how GPT models perform across diverse demographic groups and how potential biases can be mitigated.

    \item [\ding{113}] \textbf{Real-world Applications:} Assess the application of these models in real-world scenarios across different industries, such as healthcare, finance, retail, etc.
    
    \item [\ding{113}] \textbf{Interpretability and Explainability:} Focus on developing techniques to understand better and explain the predictions made by GPT models in sentiment analysis.
\end{itemize}

\balance
\bibliographystyle{IEEEtran}
\bibliography{References}

\end{document}